\documentclass{article}

\usepackage{microtype}
\usepackage{graphicx}
\usepackage{subcaption}
\usepackage{booktabs} 

\usepackage{multirow}
\usepackage[table]{xcolor} 

\usepackage{hyperref}

\usepackage[preprint]{icml2026}

\usepackage{amsmath}
\usepackage{amssymb}
\usepackage{mathtools}
\usepackage{amsthm}

\usepackage[capitalize,noabbrev]{cleveref}

\theoremstyle{plain}

\theoremstyle{definition}

\theoremstyle{remark}

\usepackage[disable]{todonotes}

\icmltitlerunning{A Unified Candidate Set with Scene-Adaptive Refinement via Diffusion for End-to-End Autonomous Driving}

\begin{document}

\twocolumn[
  \icmltitle{A Unified Candidate Set with Scene-Adaptive Refinement via Diffusion for End-to-End Autonomous Driving}



  \icmlsetsymbol{equal}{*}

  \begin{icmlauthorlist}
    \icmlauthor{Zhengfei Wu}{sch}
    \icmlauthor{Shuaixi Pan}{sch}
    \icmlauthor{Shuohan Chen}{sch}
    \icmlauthor{Shuo Yang}{sch}
    \icmlauthor{Yanjun Huang}{sch}

  \end{icmlauthorlist}

  \icmlaffiliation{sch}{School of Automotive Studies, Tongji University, Shanghai, China}

  \icmlcorrespondingauthor{Shuo Yang}{\url{shuo_yang@tongji.edu.cn}}
  \icmlcorrespondingauthor{Yanjun Huang}{\url{yanjun_huang@tongji.edu.cn}}

  \icmlkeywords{Machine Learning, ICML}

  \vskip 0.3in
]



\printAffiliationsAndNotice{}  

\begin{abstract}

End-to-end autonomous driving is increasingly adopting a multimodal planning paradigm that generates multiple trajectory candidates and selects the final plan, making candidate-set design critical. A fixed trajectory vocabulary provides stable coverage in routine driving but often misses optimal solutions in complex interactions, while scene-adaptive refinement can cause \textit{over-correction} in simple scenarios by unnecessarily perturbing already strong vocabulary trajectories.
We propose \textbf{CdDrive}, which preserves the original vocabulary candidates and augments them with scene-adaptive candidates generated by vocabulary-conditioned diffusion denoising. Both candidate types are jointly scored by a shared selection module, enabling reliable performance across routine and highly interactive scenarios. We further introduce \textbf{HATNA} (Horizon-Aware Trajectory Noise Adapter) to improve the smoothness and geometric continuity of diffusion candidates via temporal smoothing and horizon-aware noise modulation. Experiments on NAVSIM v1 and NAVSIM v2 demonstrate leading performance, and ablations verify the contribution of each component.\textit{Code:} \url{https://github.com/WWW-TJ/CdDrive}.
 
\end{abstract}

\vskip -0.3in
\begin{figure}[ht]
  \begin{center}
    \centerline{\includegraphics[width=\columnwidth]{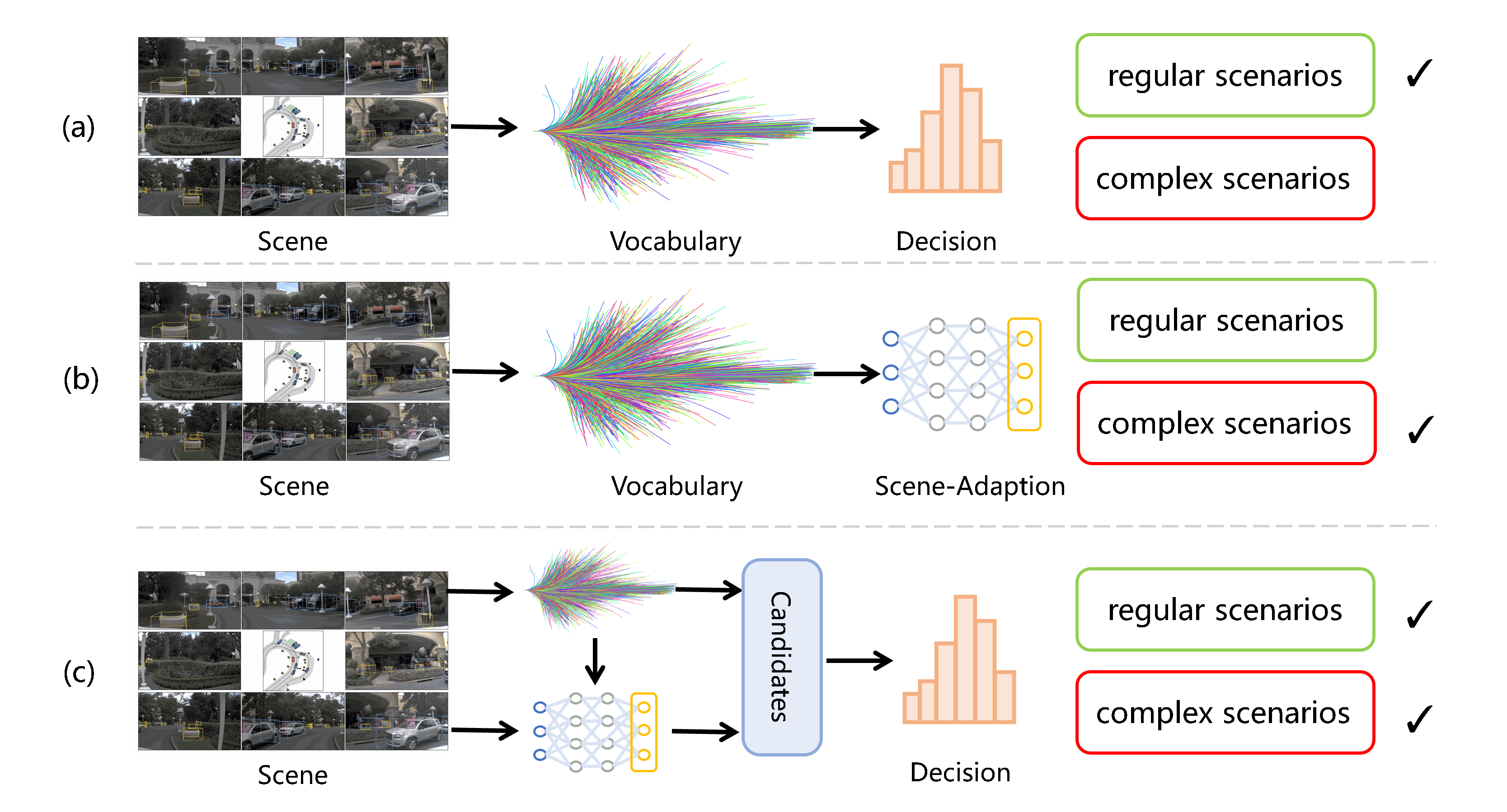}}
    \caption{
      \textbf{Candidate construction paradigms.}
      (a) Static trajectory vocabulary followed by direct online decision.
      (b) Scene-adaptive refinement on the vocabulary, followed by online decision.
      (c) \textbf{CdDrive}: unify static vocabulary candidates and diffusion-refined candidates into one candidate pool for a shared decision module.
    }
    \label{fig:iqdrive_intro}
  \end{center}
  \vskip -20pt
\end{figure}

\section{Introduction}

End-to-end autonomous driving learns a unified perception–prediction–planning model that maps multimodal sensor inputs directly to executable trajectories, reducing engineering coupling and error accumulation compared to modular pipelines \cite{liang2020pnpnet,prakash2021multi,luo2018fast,11128800}. Early planners used unimodal regression to predict a single deterministic trajectory \cite{9863660,li2025hydra,zheng2025diffusionbased}, but real driving is inherently multimodal—e.g., lane changes, merges, and unprotected intersections admit multiple reasonable futures \cite{biswas2024quad,dauner2023parting}. Thus, recent methods adopt candidate-based planning: generate multiple trajectory candidates and select the executed plan via a decision mechanism to balance diversity and safety \cite{tang2025hip,huang2025gen,11128800,xing2025goalflow}.

In this paradigm, candidate construction largely determines coverage and performance. Vocabulary-based approaches cluster offline to build fixed candidate sets and select among them \cite{chen2024vadv2,sima2025centaur,li2024hydra}, offering simple and efficient coverage for routine driving (Fig.~\ref{fig:iqdrive_intro}(a)) but limited by static support: in complex interactions the optimum may lie outside the discrete set, and enlarging the vocabulary increases computational cost. To improve adaptability, refinement methods adjust vocabulary anchors conditioned on the scene \cite{Li_2025_ICCV,Liao_2025_CVPR,zou2025diffusiondrivev2} (Fig.~\ref{fig:iqdrive_intro}(b)); e.g., WoTE applies cross-attention-based regression corrections \cite{Li_2025_ICCV,guo2025ipad}. However, when anchors are far from the optimum and interactions are complex, regression refinement can underutilize fine-grained context, leading to insufficient adaptation.

Diffusion models \cite{chi2025diffusion} provide a stronger vocabulary-guided refinement mechanism via iterative denoising: starting from a perturbed vocabulary trajectory, the model repeatedly conditions on scene context to produce a trajectory better aligned with current constraints \cite{Liao_2025_CVPR,zou2025diffusiondrivev2}. Cascaded denoising further enables coarse-to-fine refinement, and noise injection improves constraint satisfaction in complex scenes \cite{zheng2025diffusionbased,jiang2025transdiffuserdiversetrajectorygeneration}. However, a refiner trained across diverse scenes must be highly sensitive to fine-grained observations to be effective in complex interactions. This sensitivity can become a bias toward making updates even in routine driving, amplifying incidental variations and triggering \textit{over-correction}, i.e., unnecessary changes to vocabulary trajectories that degrade candidate quality.

These observations expose a key tension: (i) a static vocabulary cannot reliably include optimal solutions in complex interactive scenes, while (ii) strong scene-adaptive refinement tends to introduce redundant updates in routine scenes via over-correction. To address this, we propose \textbf{CdDrive}. CdDrive preserves static vocabulary candidates for stable routine coverage and additionally generates diffusion-refined candidates when the optimum lies beyond the vocabulary, unifying both candidate types in a shared decision module for selection (Fig.~\ref{fig:iqdrive_intro}(c)).

Unifying vocabulary and diffusion candidates also raises an overlooked issue in planning quality. Diffusion denoising injects noise into the trajectory sequence; because near- and far-horizon segments have different temporal scales, applying a uniform noise scale across timesteps can induce disproportionately large far-horizon perturbations, breaking global continuity and producing piecewise-linear artifacts. In contrast, clustered vocabulary trajectories are typically smooth. Under a mixed candidate set, the decision module may alternate between the two types across frames; if diffusion candidates are less smooth, the geometric discrepancy can introduce planning instability and degrade comfort.

To mitigate this issue, we introduce \textbf{HATNA} (Horizon-Aware Trajectory Noise Adapter), which adapts diffusion noising via temporal smoothing and horizon-aware scale modulation. HATNA suppresses the near-/far-horizon imbalance induced by independent noise, reducing kinks and improving the structural continuity and smoothness of diffusion candidates, making them more consistent with vocabulary candidates under the mixed-candidate setting of CdDrive.

We benchmark \textbf{CdDrive} on NAVSIM v1 \cite{dauner2024navsim} and NAVSIM v2 \cite{cao2025pseudo} under the standard evaluation protocol. Results demonstrate leading performance on both benchmarks, and ablations validate the contribution of each component. Our main contributions are:

\begin{itemize}
\item\textbf{Diffusion refinement for complex interactive scenarios.} We replace cross-attention-based regression refinement with diffusion-style denoising refinement, enabling richer context-conditioned interaction and exploration, and improving candidate feasibility and effectiveness in complex interactions.
\item\textbf{The CdDrive planning framework.} CdDrive combines a static trajectory vocabulary with diffusion-refined scene-adaptive candidates, and selects among them using a shared trajectory decision module to achieve robust performance across both routine and highly interactive scenarios.
\item\textbf{The HATNA noise adaptation module.} To address far-horizon kinks caused by scale inconsistency during diffusion noising, HATNA improves structural continuity through temporal smoothing and horizon-aware scaling, producing diffusion candidates that are geometrically more consistent with vocabulary candidates and improving stability under mixed candidates.
\item\textbf{Extensive evaluation on NAVSIM v1 and v2.} We conduct comprehensive experiments with ablations and qualitative analyses to demonstrate the effectiveness of CdDrive and HATNA across both routine and highly interactive scenarios.
\end{itemize}

\section{Related Works}

\subsection{End-to-End Autonomous Driving}

Compared to modular pipelines that decouple perception, prediction, and planning, end-to-end approaches learn a direct mapping from multimodal observations to planned trajectories or control commands, reducing inter-module coupling and mitigating error accumulation.
Recent progress has produced several representative end-to-end frameworks. UniAD\cite{hu2023_uniad} proposes a unified architecture that generates future trajectories directly from BEV representations, demonstrating the benefits of end-to-end autonomy. VAD\cite{jiang2023vad} learns a vectorized scene representation for planning and outputs driving trajectories based on structured embeddings. Building upon this line, VADv2\cite{chen2024vadv2} introduces an offline-constructed anchor vocabulary and models planning as predicting a probability distribution over a discrete set of trajectory candidates. Following the vocabulary-candidate paradigm, Hydra-MDP\cite{li2024hydra} incorporates rule-based safety metrics into the trajectory selection process and jointly learns via multi-teacher distillation, improving selection reliability over large discrete candidate sets. DriveDPO\cite{shang2025drivedpo} further adopts a preference-optimization objective to align the policy distribution with safety preferences. ZTRS\cite{li2025ztrs} applies offline reinforcement learning to learn a selection policy over a large candidate set, pushing the upper bound of vocabulary-based planning. Despite their strong performance, these methods remain constrained by the coverage of a fixed candidate set: when the optimal trajectory in complex interactive scenes lies outside the vocabulary, the selector can only choose a suboptimal solution. Continuously scaling up the vocabulary to approximate the optimum incurs substantial computational overhead, posing challenges for real-time deployment. These advances have also motivated vocabulary-guided planning paradigms that adapt candidates to the current scene.

\subsection{Vocabulary-Guided Scene-Adaptive Trajectory Planning}

Many end-to-end trajectory planning methods learn a direct mapping from multimodal sensor observations to future planned trajectories. Such generation-from-scratch formulations often lack explicit structured priors, making it difficult to reliably capture diverse driving behaviors across heterogeneous scenarios\cite{chen2024end}. Recent planning paradigms therefore increasingly leverage a trajectory vocabulary as a prior and perform scene-adaptive refinement on top of vocabulary anchors. WoTE\cite{Li_2025_ICCV} interacts with the current BEV state via cross-attention using trajectory-vocabulary queries, and regresses scene-conditioned offsets with an MLP. iPad\cite{guo2025ipad} similarly employs cross-attention conditioned on extracted image features to iteratively refine trajectory proposals.

More recently, diffusion models have been explored as an effective mechanism for scene-adaptive refinement due to their strong generative capacity. DiffusionDrive\cite{Liao_2025_CVPR} proposes an anchor-based truncated diffusion strategy, achieving competitive results with only a small number of denoising steps. ResAD\cite{zheng2025resad} formulates planning as residual prediction on top of an inertial reference trajectory, and combines residual normalization with reference perturbation to generate diverse candidates that better match scene constraints. DiffRefiner\cite{yin2025diffrefiner} adopts a coarse-to-fine two-stage refinement pipeline: it first regresses coarse candidates from predefined anchors, and then applies a diffusion refiner to iteratively denoise and improve them. Despite their gains in complex scenarios, refinement-based approaches can introduce unnecessary and redundant changes in simple, routine driving, resulting in \textit{over-correction}, where refinement still makes unnecessary modifications to vocabulary trajectories and degrades candidate quality.

\begin{figure*}[t]
  \vskip 0.2in
  \begin{center}
    \centerline{\includegraphics[width=\textwidth]{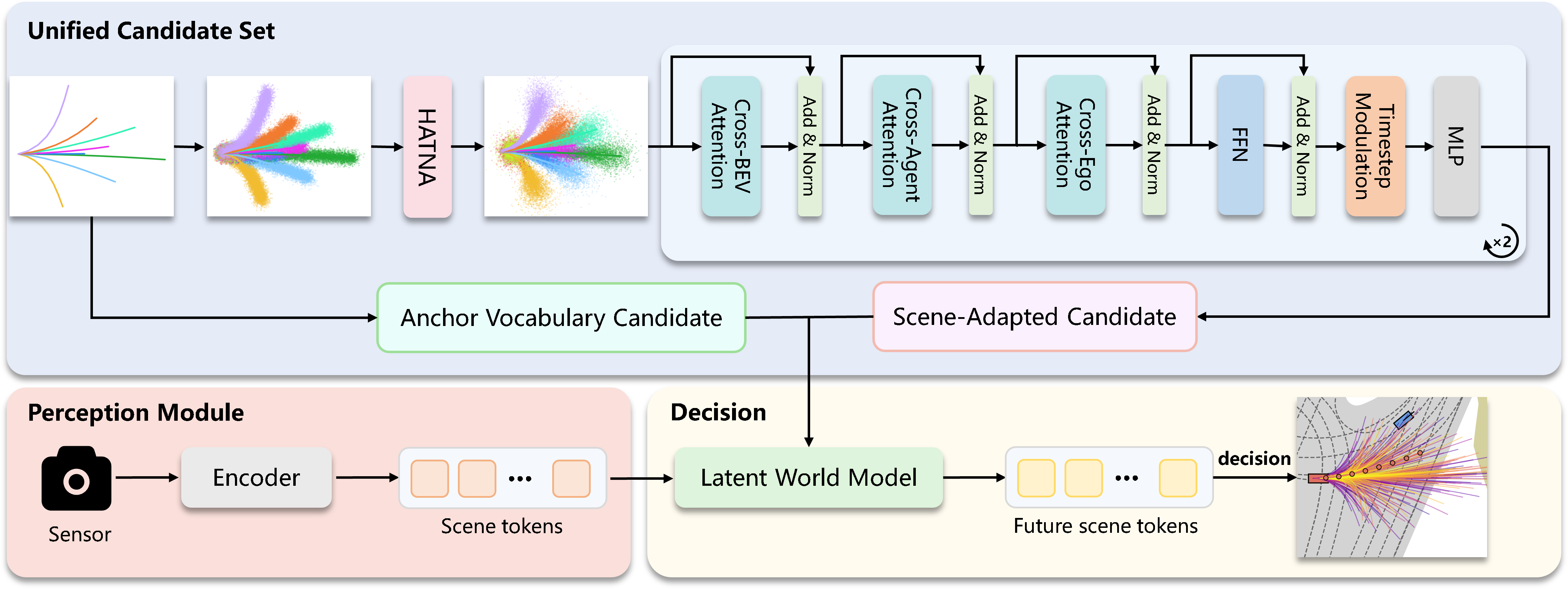}}
    \caption{
      \textbf{Overview of CdDrive.}
      CdDrive forms a unified candidate set by combining vocabulary-anchor candidates with diffusion-refined, scene-adaptive candidates.
      HATNA performs horizon-aware noise adaptation prior to denoising refinement to improve geometric continuity and smoothness.
      A shared trajectory decision module evaluates all candidates and selects the executed trajectory.
    }
    \label{fig:overall}
  \end{center}
  \vskip -0.2in
\end{figure*}

\section{Method}
We propose CdDrive, whose overall architecture is shown in Fig.~\ref{fig:overall}.
The key idea is to leverage a predefined trajectory vocabulary to provide stable candidate coverage for routine driving, while introducing diffusion-based refinement conditioned on vocabulary anchors to improve candidate quality in complex interactive scenes.
As illustrated in Fig.~\ref{fig:overall}, CdDrive unifies the original vocabulary candidates and the diffusion-refined candidates, and feeds them into a shared trajectory decision module for evaluation and selection.
Moreover, we introduce HATNA (Horizon-Aware Trajectory Noise Adapter) to adapt the injected noise in a horizon-aware manner, making diffusion refinement better respect temporal smoothness and thereby improving planning stability and comfort.

\subsection{Preliminary}

\paragraph{End-to-End Trajectory Planning.}
End-to-end autonomous driving planning aims to learn a mapping from multimodal observations to the future trajectory of the ego-vehicle.
Given the current observation $z$, where $z$ denotes the scene representation encoded by the perception backbone from multimodal sensor observations, the planner outputs a future trajectory:
\begin{equation}
    \tau = \{(x_i, y_i, \psi_i)\}_{i=1}^n,
    \label{eq:traj}
\end{equation}
where $(x_i, y_i)$ represents the planar position, $\psi_i$ is the heading angle, and $n$ is the number of discrete points within the planning horizon $H$.
We denote the 2D position subsequence as
\begin{equation}
    \mathbf{p} = \{(x_i, y_i)\}_{i=1}^n.
    \label{eq:pos_seq}
\end{equation}

\paragraph{Trajectory Vocabulary.}
We extract trajectories from expert demonstrations and utilize K-Means clustering to obtain a discrete trajectory vocabulary
\begin{equation}
\mathcal{V} = \{\tau^k\}_{k=1}^K, \quad
\tau^k = \{(x_i^k, y_i^k, \psi_i^k)\}_{i=1}^n.
\label{eq:voc}
\end{equation}

\paragraph{Truncated Conditional Diffusion.}
We establish a conditional diffusion model for refining the 2D position sequence $\mathbf{p}$ conditioned on the observation $z$.
Given a noise schedule $\{\alpha_t\}_{t=1}^T$, the forward noising process is defined as
\begin{equation}
q(\mathbf{p}_t \mid \mathbf{p}_{t-1}) =
\mathcal{N}\!\left(\sqrt{\alpha_t}\,\mathbf{p}_{t-1}, \, (1-\alpha_t)\mathbf{I}\right),
\label{eq:add_noise_single}
\end{equation}
where $\mathcal{N}(\mu,\Sigma)$ denotes a multivariate Gaussian distribution with mean $\mu$ and covariance $\Sigma$, and $\mathbf{I}$ is an identity matrix with the same dimension as $\mathbf{p}$.
By induction, the closed-form expression is
\begin{equation}
\begin{aligned}
\mathbf{p}_t &= \sqrt{\bar{\alpha}_t}\,\mathbf{p}_0
+ \sqrt{1-\bar{\alpha}_t}\,\epsilon,\\
\epsilon &\sim \mathcal{N}(\mathbf{0}, \mathbf{I}),
\bar{\alpha}_t = \prod_{s=1}^{t} \alpha_s.
\end{aligned}
\label{eq:forward_process}
\end{equation}

therefore
\begin{equation}
q(\mathbf{p}_t \mid \mathbf{p}_0) =
\mathcal{N}\!\left(\sqrt{\bar{\alpha}_t}\,\mathbf{p}_0, \, (1-\bar{\alpha}_t)\mathbf{I}\right).
\end{equation}

In the denoising phase, the model learns the reverse process conditioned on the observation:
\begin{equation}
p_\theta(\mathbf{p}_{t-1} \mid \mathbf{p}_t, z)
= \mathcal{N}\!\left(\mu_\theta(\mathbf{p}_t, z, t),\, \Sigma_t\right).
\label{eq:reverse_process}
\end{equation}
We run the reverse process for $t=T,\ldots,1$,where $\Sigma_t$ follows the standard diffusion variance schedule.
We adopt an \emph{anchor-refinement} parameterization for $\mu_\theta$.
Specifically, given an anchor trajectory $\mathbf{p}^k$ from the vocabulary, the denoising network predicts a refinement
$\Delta \mathbf{p}_\theta(\mathbf{p}_t^k, z, t)$, yielding an estimate of the refined trajectory
\begin{equation}
\hat{\mathbf{p}}_{\theta}^{k} = \mathbf{p}^{k} + \Delta \mathbf{p}_\theta(\mathbf{p}_t^{k}, z, t).
\label{eq:refine}
\end{equation}
The reverse mean $\mu_\theta(\mathbf{p}_t^{k}, z, t)$ is then obtained in closed form from $\hat{\mathbf{p}}_{\theta}^{k}$ and $\mathbf{p}_t^{k}$ according to the standard DDIM\cite{song2021denoising} update rule at step $t$.

To improve inference efficiency, we adopt a \emph{truncated} diffusion schedule\cite{Liao_2025_CVPR}:
instead of denoising from pure noise $\mathbf{p}_T \sim \mathcal{N}(0,\mathbf{I})$,
we initialize the reverse process from a noised anchor at an intermediate step $t_{tr}\ll T$, i.e., $\mathbf{p}^{k}_{t_{tr}}$,
and only perform denoising for $t=t_{tr},\ldots,1$.

\subsection{HATNA: Horizon-Aware Trajectory Noise Adapter}

We perform diffusion on the position-sequence space $\mathbf{p}=\{(x_i,y_i)\}_{i=1}^{n}$.
Standard practice samples i.i.d.\ noise at each timestep according to~\cref{eq:forward_process}.
However, due to the inconsistency in temporal and spatial scales between the near and far ends of a trajectory, applying noise with the same scale at every timestep can introduce ``zig-zag'' artifacts along the temporal dimension.
To address this, we propose HATNA, which adapts noise via an explicit function
\begin{equation}
\epsilon' = \mathrm{HATNA}(\epsilon),
\end{equation}
and replaces only the noise-addition term:
\begin{equation}
\mathbf{p}_t = \sqrt{\bar{\alpha}_t}\,\mathbf{p}_0 + \sqrt{1-\bar{\alpha}_t}\,\epsilon',
\quad \epsilon'=\mathrm{HATNA}(\epsilon).
\label{eq:hatna_noise}
\end{equation}
HATNA consists of two components: low-pass smoothing along the temporal dimension and horizon-aware scale modulation:
\begin{equation}
\mathrm{HATNA}(\epsilon) = \mathcal{S}(\epsilon)\odot \mathbf{s},
\end{equation}
where $\mathcal{S}(\cdot)$ is a temporal low-pass smoothing operator, $\mathbf{s}$ is a horizon-aware scale vector (smaller at the near end and larger at the far end), and $\odot$ denotes element-wise scaling at each timestep.
HATNA operates only on the injected noise and does not change the diffusion schedule or the denoising network, incurring negligible computational overhead.See details in\ref{app:hatna_details}\&\ref{app:hatna_hparams}.

\begin{table*}[t]
  \caption{\textbf{Performance on NAVSIM v1 with a ResNet-34 backbone.} Best results are highlighted in bold.}
  \label{tab:navsim_v1_main}
  \centering
  \begin{small}
  \setlength{\tabcolsep}{6pt}
  \renewcommand{\arraystretch}{1.05}
  \begin{tabular}{l|c|ccccc>{\columncolor{gray!20}}c}
    \toprule
    Method & Img.\ Backbone & \textbf{NC}$\uparrow$ &
    \textbf{DAC}$\uparrow$ &
    \textbf{EP}$\uparrow$ &
    \textbf{TTC}$\uparrow$ &
    \textbf{Comf.}$\uparrow$ &
    \textbf{PDMS}$\uparrow$ \\
    \midrule
    VADv2\cite{chen2024vadv2}          & ResNet-34 & 97.2 & 89.1 & 76.0 & 91.6 & \textbf{100.0} & 80.9 \\
    UniAD\cite{hu2023_uniad}          & ResNet-34 & 97.8 & 91.9 & 78.8 & 92.9 & \textbf{100.0} & 83.4 \\
    PARA-Drive\cite{weng2024drive}     & ResNet-34 & 97.9 & 92.4 & 79.3 & 93.0 & 99.8          & 84.0 \\
    Transfuser\cite{9863660}     & ResNet-34 & 97.7 & 92.8 & 79.2 & 92.8 & \textbf{100.0} & 84.0 \\
    LAW\cite{li2025enhancing}            & ResNet-34 & 96.4 & 95.4 & 81.7 & 88.7 & 99.9          & 84.6 \\
    DRAMA\cite{yuan2024drama}          & ResNet-34 & 98.0 & 93.1 & 80.1 & 94.8 & \textbf{100.0} & 85.5 \\
    GoalFlow\cite{xing2025goalflow}       & ResNet-34 & 98.3 & 93.8 & 79.8 & 94.3 & \textbf{100.0} & 85.7 \\
    Hydra-MDP\cite{li2024hydra}      & ResNet-34 & 98.3 & 96.0 & 78.7 & 94.6 & \textbf{100.0} & 86.5 \\
    ARTEMIS\cite{feng2025artemis}        & ResNet-34 & 98.3 & 95.1 & 81.4 & 94.3 & \textbf{100.0} & 87.0 \\
    DiffusionDrive\cite{Liao_2025_CVPR} & ResNet-34 & 98.2 & 96.2 & 82.2 & 94.7 & \textbf{100.0} & 88.1 \\
    WoTE\cite{Li_2025_ICCV}           & ResNet-34 & 98.5 & 96.8 & 81.9 & 94.9 & 99.9          & 88.3 \\
    DRIVER\cite{song2025breaking}         & ResNet-34 & 98.5 & 96.5 & 82.6 & 94.9 & \textbf{100.0} & 88.3 \\
    ResAD\cite{zheng2025resad}          & ResNet-34 & 98.0 & 97.3 & 82.5 & 94.2 & \textbf{100.0} & 88.6 \\
    \midrule
    Ours           & ResNet-34 & \textbf{98.7} & \textbf{97.5} & \textbf{82.7} & \textbf{95.4} & \textbf{100.0} & \textbf{89.2} \\
    \bottomrule
  \end{tabular}
  \end{small}
  \vskip -0.1in
\end{table*}
\subsection{Scene-Adapted Candidate Generation with Diffusion}

We now introduce how CdDrive generates scene-adapted candidates via prior-guided truncated diffusion on top of the trajectory vocabulary.
Given the vocabulary $\mathcal{V}=\{\tau^{k}\}_{k=1}^{K}$, we take the position subsequence of each anchor trajectory as
$\mathbf{p}^{k}=\{(x_i^{k},y_i^{k})\}_{i=1}^{n}$,
and treat $\mathbf{p}^{k}$ as a strong prior for anchor-refinement conditioned on the observation $z$.

\paragraph{Truncated noising initialization.}
For each anchor $\mathbf{p}^{k}$, we first sample base noise $\epsilon\sim\mathcal{N}(0,\mathbf{I})$ and adapt it via HATNA as $\epsilon'=\mathrm{HATNA}(\epsilon)$.
Then, using the closed-form forward process in~\cref{eq:hatna_noise}, we initialize the reverse process from a noised anchor at an intermediate timestep $t_{tr}\ll T$:
\begin{equation}
\mathbf{p}^{k}_{t_{tr}}
= \sqrt{\bar{\alpha}_{t_{tr}}}\,\mathbf{p}^{k} + \sqrt{1-\bar{\alpha}_{t_{tr}}}\,\epsilon'.
\end{equation}
This truncated schedule avoids denoising from pure noise and reduces the number of denoising steps.

\paragraph{Prior-guided denoising with anchor-refinement.}
Starting from $\mathbf{p}^{k}_{t_{tr}}$, we run a conditional denoising decoder with DDIM sampling and stride $\Delta$ for
$t=t_{tr},t_{tr}-\Delta,\ldots,1$.
At each step $t$, the network predicts an anchor-refinement $\Delta \mathbf{p}_\theta(\mathbf{p}_t^{k}, z, t)$ and computes the refined-trajectory estimate following~\cref{eq:refine},
from which the reverse mean $\mu_{\theta}(\mathbf{p}^{k}_{t},z,t)$ is obtained in closed form according to the standard DDIM update rule.
After the final denoising step (at $t=0$), we obtain the scene-adapted position sequence:
\begin{equation}
\tilde{\mathbf{p}}^{k} \triangleq \hat{\mathbf{p}}_{\theta}^{k}(t{=}0).
\end{equation}

\paragraph{Heading prediction.}
Due to the inherent $2\pi$ periodicity of the heading angle, directly injecting Gaussian diffusion noise in $\psi$ can introduce discontinuities at the wrap-around boundary and destabilize training.
Therefore, we do not inject diffusion noise in the heading dimension.
Instead, the denoising decoder outputs the heading sequence through an additional MLP head, and we constrain it to a valid range using $\pi\tanh(\cdot)$:
\begin{equation}
\tilde{\psi}^{k}_{i} = \pi \tanh\!\big(g_{\theta}(\mathbf{p}^{k}_{t},z,t)_i\big), \quad i=1,\ldots,n,
\end{equation}
where $g_\theta(\cdot)$ shares the same denoising decoder with the position-refinement head and differs only in the final output head.
This design preserves stable learning while keeping heading prediction consistent with the refined positions.
We then construct the refined trajectory candidate as
\begin{equation}
\tilde{\tau}^{k}=\{(\tilde{x}^{k}_{i},\tilde{y}^{k}_{i},\tilde{\psi}^{k}_{i})\}_{i=1}^{n}.
\end{equation}

\paragraph{Scene-adapted candidate set.}
Repeating the above process for all anchors yields a diffusion-refined candidate set
$\tilde{\mathcal{V}}(z)=\{\tilde{\tau}^{k}\}_{k=1}^{K}$.
These candidates complement the static vocabulary by providing scene-adapted solutions in complex interactions.

\subsection{Trajectory Decision}
\label{sec:trajectory_decision}
\paragraph{A unified candidate set.}
Given the observation $z$, CdDrive produces a unified candidate set by combining the static vocabulary and the diffusion-refined candidates:
\begin{equation}
\mathcal{C}(z) \triangleq \mathcal{V} \cup \tilde{\mathcal{V}}(z)
= \{\tau^{k}\}_{k=1}^{K} \cup \{\tilde{\tau}^{k}\}_{k=1}^{K}.
\end{equation}
The goal of trajectory decision is to select the final executed trajectory $\tau^*$ from $\mathcal{C}(z)$.

\paragraph{Latent world model rollout.}
Our trajectory decision process follows WoTE\cite{Li_2025_ICCV}.Before scoring, we leverage a latent world model\cite{li2025enhancing} to predict future scene evolution under each candidate trajectory.
Given the current observation $z$ and a candidate $\tau\in\mathcal{C}(z)$, the latent world model rolls out future observations in a latent feature space following the planned motion:
\begin{equation}
\tilde{z} = \mathrm{LWM}(\tau, z).
\end{equation}
The latent world model is trained jointly with the planner and the decision module.See details in\ref{app:lwm_supervision}.

\paragraph{Scene-conditioned scoring.}
We employ a trajectory decision module, a scene-conditioned scorer $f_\phi(\cdot)$ that maps each candidate trajectory, the current observation, and the world-model prediction to a vector of scores:
\begin{equation}
\mathbf{s}(\tau; z) = f_\phi(\tau, z, \tilde{z}),
\end{equation}
where each component of $\mathbf{s}$ corresponds to one scoring term that captures different planning objectives, including safety-, progress-, and comfort-related scores, and a higher value indicates a more preferable trajectory under the current scene context.
Importantly, the scorer evaluates \emph{both} the original vocabulary candidates and the diffusion-refined candidates in the same scoring space, enabling direct comparison across the two sources of proposals.

\paragraph{Decision rule.}
Following\cite{li2024hydra,Li_2025_ICCV,yao2025drivesuprim},at inference time, we aggregate the score vector via a weighted sum to obtain an overall score:
\begin{equation}
S(\tau; z) = \mathbf{w}^\top \mathbf{s}(\tau; z),
\end{equation}
where $\mathbf{w}$ is a fixed weight vector pre-specified for different scoring terms. We then select the executed trajectory as
\begin{equation}
\tau^* = \arg\max_{\tau \in \mathcal{C}(z)} S(\tau; z),
\end{equation}
selecting the candidate with the highest aggregated score.

\subsection{Learning Objectives}
\label{sec:learning_objectives}

We train CdDrive to jointly optimize two core components:
(i) the diffusion-based anchor-refinement denoising decoder that generates scene-adapted trajectory candidates, and
(ii) the trajectory decision module that scores candidates with multiple criteria.
The overall objective is a weighted combination of the following loss terms.See details in\ref{app:aux_losses}

\paragraph{Anchor-refinement supervision.}
Let $\boldsymbol{\tau}^{gt}$ denote the expert trajectory for the current scene.
The denoising decoder refines anchor trajectories to produce scene-adapted candidates.
We apply WTA(winner-take-all) supervision\cite{zhou2023query} on the denoised trajectories by selecting the candidate with the smallest $\ell_2$ distance to $\boldsymbol{\tau}^{gt}$ as the winner, denoted by $\boldsymbol{\tau}_{\text{winner}}$.
We supervise the winner using an $\ell_1$ loss:
\begin{equation}
\mathcal{L}_{traj}
= \left\lVert \boldsymbol{\tau}_{\text{winner}} - \boldsymbol{\tau}^{gt} \right\rVert_{1}.
\end{equation}

\paragraph{Imitation scoring loss.}
The decision module predicts an imitation-related score distribution over anchors.
We construct a soft target distribution based on the anchor-to-expert distance:
\begin{equation}
r_{im}^k
= \frac{\exp\!\big(-\lVert \boldsymbol{\tau}^{k}-\boldsymbol{\tau}^{gt}\rVert_2\big)}
{\sum_{j=1}^{K}\exp\!\big(-\lVert \boldsymbol{\tau}^{j}-\boldsymbol{\tau}^{gt}\rVert_2\big)} .
\end{equation}
Let $\mathbf{r}_{im}=\{r_{im}^k\}_{k=1}^{K}$ denote the soft target distribution and
$\hat{\mathbf{r}}_{im}=\{\hat{r}_{im}^k\}_{k=1}^{K}$ the prediction.
We minimize the cross-entropy:
\begin{equation} 
\mathcal{L}_{im}=\mathrm{CE}\!\left(\hat{\mathbf{r}}_{im},\,\mathbf{r}_{im}\right).
\end{equation}

\paragraph{Simulation-based scoring loss.}
We train simulation-related scoring terms using precomputed targets.
Let $r_{sim}^k\in[0,1]$ be the target score and $\hat{r}_{sim}^k\in(0,1)$ the prediction for the $k$-th candidate.
We use the binary cross-entropy loss:
\begin{equation}
\mathcal{L}_{sim}=\frac{1}{K}\sum_{k=1}^{K}\mathrm{BCE}\!\left(\hat{r}_{sim}^k,\, r_{sim}^k\right).
\label{eq:lsim}
\end{equation}

\paragraph{Overall objective.}
The final training objective is
\begin{equation}
\mathcal{L}
= \lambda_{traj}\mathcal{L}_{traj}
+ \lambda_{im}\mathcal{L}_{im}
+ \lambda_{sim}\mathcal{L}_{sim},
\end{equation}
where $\lambda_{traj}$, $\lambda_{im}$, and $\lambda_{sim}$ are fixed weights.

\begin{table*}[t]
  \caption{\textbf{Performance on NAVSIM v2.} Best results are highlighted in bold.}
  \label{tab:navsim_v2_main}
  \centering
  \begin{small}
  \setlength{\tabcolsep}{6pt}
  \renewcommand{\arraystretch}{1.05}
  \begin{tabular}{lccccccccc>{\columncolor{gray!20}}c}
    \toprule
    Method & \textbf{NC}$\uparrow$ &
    \textbf{DAC}$\uparrow$ &
    \textbf{DDC}$\uparrow$ &
    \textbf{TLC}$\uparrow$ &
    \textbf{EP}$\uparrow$ &
    \textbf{TTC}$\uparrow$ &
    \textbf{LK}$\uparrow$ &
    \textbf{HC}$\uparrow$ &
    \textbf{EC}$\uparrow$ &
    \textbf{EPDMS}$\uparrow$ \\
    \midrule
    Ego Status MLP & 93.1 & 77.9 & 92.7 & 99.6 & 86.0 & 91.5 & 89.4 & 98.3 & 85.4 & 64.0 \\
    Transfuser\cite{9863660}     & 96.9 & 89.9 & 97.8 & 99.7 & 87.1 & 95.4 & 92.7 & 98.3 & 87.2 & 76.7 \\
    HydraMDP++\cite{li2025hydra}     & 97.2 & \textbf{97.5} & 99.4 & 99.6 & 83.1 & 96.5 & 94.4 & 98.2 & 70.9 & 81.4 \\
    DriveSuprim\cite{yao2025drivesuprim}    & 97.5 & 96.5 & 99.4 & 99.6 & \textbf{88.4} & 96.6 & 95.5 & 98.3 & 77.0 & 83.1 \\
    ARTEMIS\cite{feng2025artemis}        & \textbf{98.3} & 95.1 & 98.6 & \textbf{99.8} & 81.5 & 97.4 & 96.5 & 98.3 & --   & 83.1 \\
    DiffusionDrive\cite{Liao_2025_CVPR} & 98.2 & 95.9 & 99.4 & \textbf{99.8} & 87.5 & 97.3 & 96.8 & 98.3 & 87.7 & 84.5 \\
    ResAD\cite{zheng2025resad}          & 97.8 & 97.2 & 99.5 & \textbf{99.8} & 88.2 & 96.9 & 97.0 & \textbf{98.4} & \textbf{88.2} & 85.5 \\
    \midrule
    Ours           & \textbf{98.3} & 97.4 & \textbf{99.6} & \textbf{99.8} & \textbf{88.4} & \textbf{97.9} & \textbf{97.5} & \textbf{98.4} & \textbf{88.2} & \textbf{86.4} \\
    \bottomrule
  \end{tabular}
  \end{small}
  \vskip -0.1in
\end{table*}

\section{Experiments}

We evaluate \textbf{CdDrive} on the NAVSIM v1\cite{dauner2024navsim} and NAVSIM v2\cite{cao2025pseudo} benchmarks, which are constructed from nuPlan\cite{caesar2021nuplan} real-world driving logs and built on top of the OpenScene\cite{contributors2023openscene} preprocessing pipeline. OpenScene downsamples raw nuPlan data from $10\,\mathrm{Hz}$ to $2\,\mathrm{Hz}$, and NAVSIM further re-samples these frames to emphasize challenging segments with dynamic intent changes while filtering out overly trivial portions (e.g., long stationary or constant-speed driving). Each frame provides $8$ surround-view high-resolution camera images and fused point clouds from $5$ LiDAR sensors. The dataset is split into \texttt{navtrain} and \texttt{navtest}; \texttt{navtrain} contains $1192$ scenarios for training/validation and \texttt{navtest} contains $136$ scenarios for testing.

\subsection{Benchmark}

\paragraph{NAVSIM v1.}
The primary metric in NAVSIM v1\cite{dauner2024navsim} is the Predictive Driver Model Score (PDMS), which aggregates five sub-metrics: NC (No At-Fault Collision), DAC (Drivable Area Compliance), EP (Ego Progress), TTC (Time-to-Collision), and Comf (Comfort). Following the benchmark definition, PDMS is computed as
\begin{equation}
\label{eq:pdms}
\mathrm{PDMS}
=
\mathrm{NC}\cdot \mathrm{DAC}\cdot
\frac{5\,\mathrm{EP}+5\,\mathrm{TTC}+2\,\mathrm{Comf}}{12}.
\end{equation}

\paragraph{NAVSIM v2.}
NAVSIM v2\cite{cao2025pseudo} extends PDMS to the Extended PDM Score (EPDMS) by introducing additional compliance and comfort terms. Specifically, it adds DDC (Driving Direction Compliance), TLC (Traffic Light Compliance), and LK (Lane Keeping), and replaces Comf with HC (History Comfort) and EC (Extended Comfort). EPDMS is computed as
\begin{equation}
\label{eq:epdms}
\begin{aligned}
\mathrm{EPDMS}
=
\mathrm{NC}\cdot \mathrm{DAC}\cdot \mathrm{DDC}\cdot \mathrm{TLC}\cdot \frac{1}{16}
\\
\quad \left(
5\,\mathrm{EP}+5\,\mathrm{TTC}+2\,\mathrm{LK}+2\,\mathrm{HC}+2\,\mathrm{EC}
\right).
\end{aligned}
\end{equation}

\subsection{Implementation Details}

We follow Transfuser\cite{9863660} for perception. The input consists of three forward-facing camera views that are cropped and downsampled, then concatenated into a $1024\times256$ image tensor, together with a LiDAR point cloud covering a $64\,\mathrm{m}\times64\,\mathrm{m}$ region centered at the ego vehicle. The perception module encodes these inputs into BEV features, and we use ResNet-34\cite{he2016deep} as the image backbone for fair comparison. For planning, \textbf{CdDrive} uses a trajectory vocabulary of size $K=256$ and a two-stage cascaded denoising decoder. We train on \texttt{navtrain} for 30 epochs using AdamW with a learning rate of $10^{-4}$ and a weight decay of $0.01$, together with a 5-epoch warmup and cosine annealing.The total batch size is $128$ across $8$ GPUs.See details in\ref{app:exp_details}

\subsection{Main Results}

\paragraph{NAVSIM v1.}
Table~\ref{tab:navsim_v1_main} reports the NAVSIM v1 results with a ResNet-34 backbone. \textbf{CdDrive} achieves the best overall PDMS of \textbf{89.2}, outperforming both vocabulary-based planners and recent diffusion-based methods. \textbf{CdDrive} also attains the best performance on the key sub-metrics \textbf{NC}, \textbf{DAC}, \textbf{EP}, and \textbf{TTC}, which verifies the effectiveness of our unified-candidate formulation that retains the original vocabulary trajectories for routine scenes while augmenting them with diffusion-refined, scene-adapted candidates for complex interactions where the optimum can lie outside the static set. Meanwhile, \textbf{CdDrive} maintains \textbf{100.0} comfort, This further corroborates the effectiveness of \textbf{HATNA} in improving the overall smoothness of diffusion candidates.

\paragraph{NAVSIM v2.}
Table~\ref{tab:navsim_v2_main} summarizes the NAVSIM v2 results. \textbf{CdDrive} achieves the best overall EPDMS of \textbf{86.4} and maintains consistently strong performance across the extended compliance and comfort metrics. These results suggest that the unified candidate set and horizon-aware noise adaptation remain effective under NAVSIM v2's more challenging extended metric suite.


\begin{table}[t]
  \caption{Ablation study on candidate set design on NAVSIM v1. Best results are highlighted in bold.}
  \label{tab:ablation_candidates}
  \centering
  \begin{small}
  \setlength{\tabcolsep}{3.0pt}
  \renewcommand{\arraystretch}{1.05}
  \resizebox{\columnwidth}{!}{%
  \begin{tabular}{l c c c c c c >{\columncolor{gray!20}}c}
    \toprule
    Method & \#Traj. &
    \textbf{NC}$\uparrow$ &
    \textbf{DAC}$\uparrow$ &
    \textbf{EP}$\uparrow$ &
    \textbf{TTC}$\uparrow$ &
    \textbf{Comf}$\uparrow$ &
    \cellcolor{gray!15}\textbf{PDMS}$\uparrow$ \\
    \midrule
    Vocab-only     & 256      & 98.2 & 96.3 & 81.0 & 94.3 & 100.0 & 87.4 \\
    Vocab-only     & 512      & 98.4 & 95.9 & 81.0 & 94.7 & 100.0 & 87.4 \\
    Diffusion-only & 256      & 98.5 & 97.3 & 82.5 & 94.4 & 100.0 & 88.6 \\
    Diffusion-only & 512      & 98.6 & 97.1 & 82.1 & 94.7 & 100.0 & 88.4 \\
    \midrule
    \textbf{CdDrive}        & 256+256  & \textbf{98.7} & \textbf{97.5} & \textbf{82.7} & \textbf{95.4} & \textbf{100.0} & \textbf{89.2} \\
    \bottomrule
  \end{tabular}}
  \end{small}
  \vskip -0.1in
\end{table}


\begin{table}[t]
  \caption{Ablation study of HATNA on NAVSIM v1. Best results are highlighted in bold.}
  \label{tab:ablation_hatna}
  \centering
  \begin{small}
  \setlength{\tabcolsep}{4.5pt}
  \renewcommand{\arraystretch}{1.05}
  \begin{tabular}{l c c c c c >{\columncolor{gray!20}}c}
    \toprule
    HATNA &
    \textbf{NC}$\uparrow$ &
    \textbf{DAC}$\uparrow$ &
    \textbf{EP}$\uparrow$ &
    \textbf{TTC}$\uparrow$ &
    \textbf{Comf}$\uparrow$ &
    \cellcolor{gray!15}\textbf{PDMS}$\uparrow$ \\
    \midrule
    w/o  & \textbf{98.7} & \textbf{97.5} & \textbf{82.7} & 95.0 & 99.9 & 89.0 \\
    \textbf{w/}   & \textbf{98.7} & \textbf{97.5} & \textbf{82.7} & \textbf{95.4} & \textbf{100.0} & \textbf{89.2} \\
    \bottomrule
  \end{tabular}
  \end{small}
  \vskip -0.1in
\end{table}


\begin{table}[t]
  \caption{Ablation study comparing diffusion-based and regression-based scene-adaptive refiners, and candidate set with/without the trajectory vocabulary on NAVSIM v1. Best results are highlighted in bold.}
  \label{tab:ablation_refiner}
  \centering
  \begin{small}
  \setlength{\tabcolsep}{3.5pt}
  \renewcommand{\arraystretch}{1.05}
  \resizebox{\columnwidth}{!}{%
  \begin{tabular}{l l c c c c c >{\columncolor{gray!20}}c}
    \toprule
    Refiner & Candidates &
    \textbf{NC}$\uparrow$ &
    \textbf{DAC}$\uparrow$ &
    \textbf{EP}$\uparrow$ &
    \textbf{TTC}$\uparrow$ &
    \textbf{Comf}$\uparrow$ &
    \cellcolor{gray!15}\textbf{PDMS}$\uparrow$ \\
    \midrule
    Regression & w/o Vocab & 98.5 & 96.8 & 81.9 & 94.9 & 100.0 & 88.3 \\
    Diffusion   & w/o Vocab & 98.5 & 97.3 & 82.5 & 94.4 & 100.0 & 88.6 \\
    Regression & w/  Vocab & 98.3 & 97.0 & 82.6 & 94.4 & 100.0 & 88.5 \\
    Diffusion   & w/  Vocab & \textbf{98.7} & \textbf{97.5} & \textbf{82.7} & \textbf{95.4} & \textbf{100.0} & \textbf{89.2} \\
    \bottomrule
  \end{tabular}}
  \end{small}
  \vskip -0.1in
\end{table}
\subsection{Ablation Studies}

\paragraph{Candidate set design.}
Table~\ref{tab:ablation_candidates} analyzes the impact of candidate sources and candidate set size on NAVSIM v1.
Using only the static trajectory vocabulary is consistently suboptimal, and expanding the vocabulary size from $256$ to $512$ yields no noticeable performance improvement, indicating that naive scaling of a fixed candidate set does not reliably improve planning quality.
Using only diffusion-refined candidates improves over vocab-only, yet remains limited due to the redundancy-refinement issue, where unnecessary corrections can be introduced in routine scenes and consequently degrade candidate quality.
In contrast, \textbf{CdDrive} achieves the best overall performance, supporting the motivation of unifying candidates from two paradigms: it preserves the original vocabulary trajectories to avoid redundant refinement in routine driving while augmenting them with diffusion-refined, scene-adapted candidates to better cover complex interactions where the optimum may lie outside the static set.

\paragraph{Effect of HATNA.}
Table~\ref{tab:ablation_hatna} evaluates HATNA.
Enabling HATNA improves both \textbf{TTC} and \textbf{Comf}, indicating further gains on safety- and comfort-related outcomes, and leads to a higher overall PDMS.
These results corroborate that horizon-aware noise adaptation stabilizes diffusion refinement by mitigating non-uniform perturbations along the trajectory horizon, thereby producing smoother diffusion candidates.

\paragraph{Diffusion refinement vs.\ regression refinement.}
Table~\ref{tab:ablation_refiner} compares diffusion-based refinement with regression-based scene-adaptive refinement under two candidate settings.
Diffusion-based refinement consistently outperforms regression-based refinement,
this suggests that iterative denoising is more effective at progressively aligning candidates to scene conditions than directly regressing corrections.
Compared to regression-based refinement, diffusion-based refinement yields a larger gain when combined with the trajectory vocabulary, which indicates that trajectory vocabulary and  diffusion candidates are more complementary within the unified candidate set of \textbf{CdDrive}.See more in\ref{app:analysis}

\subsection{Qualitative Analysis}

\begin{figure}[t]
  \vskip 0.2in
  \begin{center}
    \centerline{\includegraphics[width=\columnwidth]{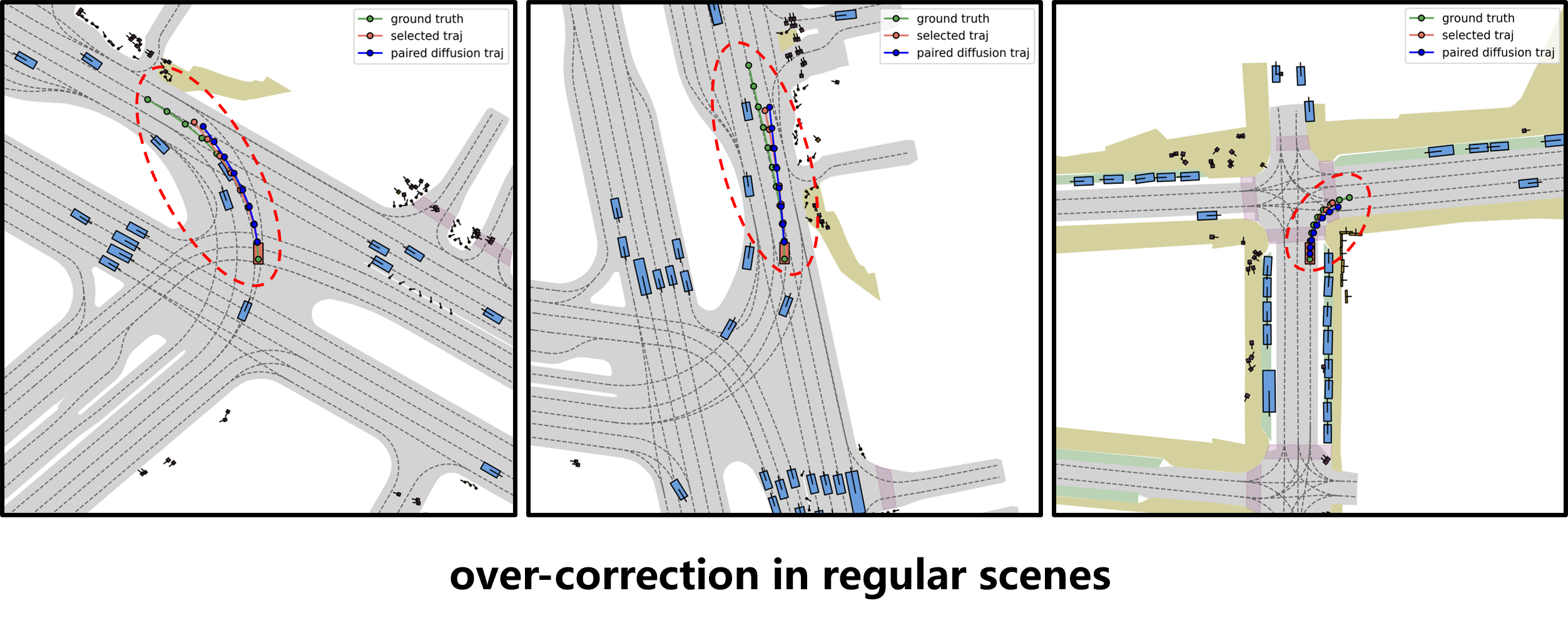}}
    \caption{
      \textbf{Qualitative examples where scene-adaptive refinement does not necessarily improve over vocabulary trajectories.}
      Green denotes the ground-truth trajectory, red denotes the selected vocabulary trajectory, and blue denotes the paired diffusion candidate.
      In routine (low-interaction) scenes, refinement can introduce \emph{over-corrections} (highlighted in red circles), where the diffusion candidate deviates from the selected vocabulary trajectory and drifts away from the ground-truth trend.
    }
  \label{fig:qual_overcorrection}
  \end{center}
  \vskip -0.3in
\end{figure}

\paragraph{Over-correction.}
Figure~\ref{fig:qual_overcorrection} illustrates a representative failure mode of refinement-based candidate generation.
In routine (low-interaction) scenes, the vocabulary already provides competitive candidates, yet diffusion refinement can still make unnecessary updates and becomes \emph{over-correction}.
In the highlighted regions, the diffusion candidate deviates from the selected vocabulary trajectory without clear benefit and can move further away from the ground-truth trend.
This phenomenon corroborates the design of \textbf{CdDrive}, which retains the original vocabulary candidates and unifies them with diffusion candidates into a single candidate pool, so that the trajectory decision module can select from the unified candidate set the trajectory that best satisfies the current scene constraints.

\begin{figure}[t]

  \begin{center}
    \centerline{\includegraphics[width=\columnwidth]{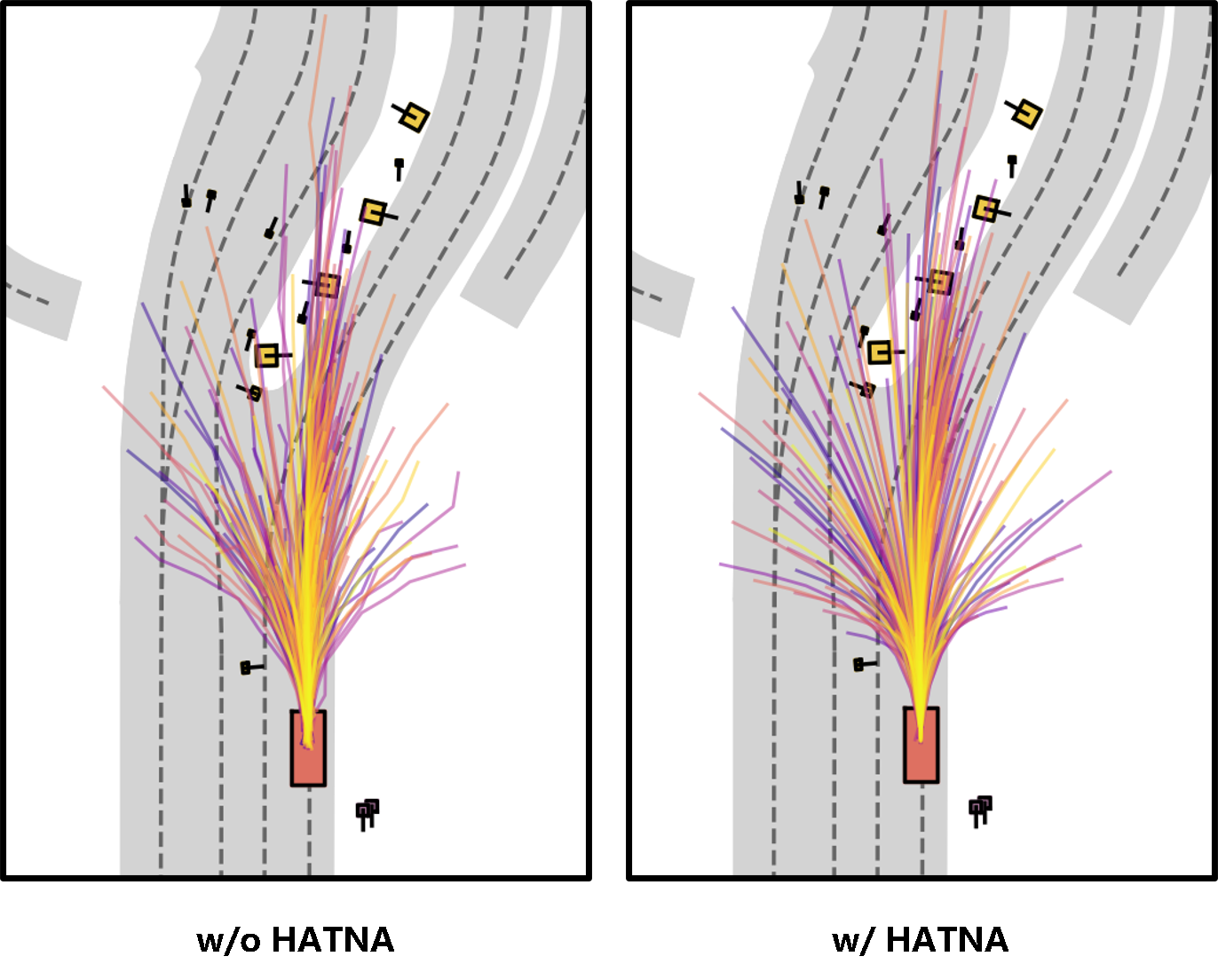}}
    \caption{
      \textbf{Qualitative comparison of diffusion candidates with and without HATNA.}
      Uniform noising across timesteps often yields kinked trajectories, while HATNA produces smoother diffusion candidates.
    }
    \label{fig:hatna_qual}
  \end{center}
  \vskip -0.5in
\end{figure}

\paragraph{HATNA.}
As shown in Fig.~\ref{fig:hatna_qual}, applying a uniform noise scale across all trajectory timesteps can lead to piecewise-linear kinks in diffusion candidates.
With HATNA, the diffusion candidate set becomes noticeably smoother and effectively suppresses such kinked patterns.
This indicates that applying temporal smoothing and horizon-aware scale modulation to the injected noise during the noising stage enables HATNA to improve the geometric continuity of diffusion candidates.

\section{Conclusion}

In this work, we propose \textbf{CdDrive}, a unified candidate-set framework that retains a predefined trajectory vocabulary and augments it with diffusion-based scene-adaptive refinements. \textbf{CdDrive} resolves a key tension in vocabulary-guided planning: static vocabularies provide stable candidates in routine driving yet lack coverage in complex interactions, while refinement-based methods can introduce over-corrections when adaptation is unnecessary. To improve the geometric quality of diffusion candidates, we further introduce HATNA, which performs temporal smoothing and horizon-aware scale modulation during the noising stage to enhance continuity and overall smoothness. Extensive experiments and qualitative analyses on NAVSIM v1 and v2 verify the effectiveness of our design choices, showing that diffusion-based refinement consistently outperforms regression-based refinement and that the unified candidate set yields robust performance across diverse driving scenarios, achieving state-of-the-art results under standard evaluation protocols.

\bibliography{example_paper}

@inproceedings{liang2020pnpnet,
  title={Pnpnet: End-to-end perception and prediction with tracking in the loop},
  author={Liang, Ming and Yang, Bin and Zeng, Wenyuan and Chen, Yun and Hu, Rui and Casas, Sergio and Urtasun, Raquel},
  booktitle={Proceedings of the IEEE/CVF Conference on Computer Vision and Pattern Recognition},
  pages={11553--11562},
  year={2020}
}

@inproceedings{prakash2021multi,
  title={Multi-modal fusion transformer for end-to-end autonomous driving},
  author={Prakash, Aditya and Chitta, Kashyap and Geiger, Andreas},
  booktitle={Proceedings of the IEEE/CVF conference on computer vision and pattern recognition},
  pages={7077--7087},
  year={2021}
}

@inproceedings{luo2018fast,
  title={Fast and furious: Real time end-to-end 3d detection, tracking and motion forecasting with a single convolutional net},
  author={Luo, Wenjie and Yang, Bin and Urtasun, Raquel},
  booktitle={Proceedings of the IEEE conference on Computer Vision and Pattern Recognition},
  pages={3569--3577},
  year={2018}
}

@INPROCEEDINGS{11128800,
  author={Sun, Wenchao and Lin, Xuewu and Shi, Yining and Zhang, Chuang and Wu, Haoran and Zheng, Sifa},
  booktitle={2025 IEEE International Conference on Robotics and Automation (ICRA)}, 
  title={SparseDrive: End-to-End Autonomous Driving via Sparse Scene Representation}, 
  year={2025},
  volume={},
  number={},
  pages={8795-8801},
  doi={10.1109/ICRA55743.2025.11128800}}

@ARTICLE{9863660,
  author={Chitta, Kashyap and Prakash, Aditya and Jaeger, Bernhard and Yu, Zehao and Renz, Katrin and Geiger, Andreas},
  journal={IEEE Transactions on Pattern Analysis and Machine Intelligence}, 
  title={TransFuser: Imitation With Transformer-Based Sensor Fusion for Autonomous Driving}, 
  year={2023},
  volume={45},
  number={11},
  pages={12878-12895},
  doi={10.1109/TPAMI.2022.3200245}}

@article{li2025hydra,
  title={Hydra-mdp++: Advancing end-to-end driving via expert-guided hydra-distillation},
  author={Li, Kailin and Li, Zhenxin and Lan, Shiyi and Xie, Yuan and Zhang, Zhizhong and Liu, Jiayi and Wu, Zuxuan and Yu, Zhiding and Alvarez, Jose M},
  journal={arXiv preprint arXiv:2503.12820},
  year={2025}
}

@inproceedings{
zheng2025diffusionbased,
title={Diffusion-Based Planning for Autonomous Driving with Flexible Guidance},
author={Yinan Zheng and Ruiming Liang and Kexin ZHENG and Jinliang Zheng and Liyuan Mao and Jianxiong Li and Weihao Gu and Rui Ai and Shengbo Eben Li and Xianyuan Zhan and Jingjing Liu},
booktitle={The Thirteenth International Conference on Learning Representations},
year={2025},
url={https://openreview.net/forum?id=wM2sfVgMDH}
}

@inproceedings{biswas2024quad,
  title={Quad: Query-based interpretable neural motion planning for autonomous driving},
  author={Biswas, Sourav and Casas, Sergio and Sykora, Quinlan and Agro, Ben and Sadat, Abbas and Urtasun, Raquel},
  booktitle={2024 IEEE International Conference on Robotics and Automation (ICRA)},
  pages={14236--14243},
  year={2024},
  organization={IEEE}
}

@inproceedings{dauner2023parting,
  title={Parting with misconceptions about learning-based vehicle motion planning},
  author={Dauner, Daniel and Hallgarten, Marcel and Geiger, Andreas and Chitta, Kashyap},
  booktitle={Conference on Robot Learning},
  pages={1268--1281},
  year={2023},
  organization={PMLR}
}

@article{tang2025hip,
  title={Hip-ad: Hierarchical and multi-granularity planning with deformable attention for autonomous driving in a single decoder},
  author={Tang, Yingqi and Xu, Zhuoran and Meng, Zhaotie and Cheng, Erkang},
  journal={arXiv preprint arXiv:2503.08612},
  year={2025}
}

@inproceedings{huang2025gen,
  title={Gen-drive: Enhancing diffusion generative driving policies with reward modeling and reinforcement learning fine-tuning},
  author={Huang, Zhiyu and Weng, Xinshuo and Igl, Maximilian and Chen, Yuxiao and Cao, Yulong and Ivanovic, Boris and Pavone, Marco and Lv, Chen},
  booktitle={2025 IEEE International Conference on Robotics and Automation (ICRA)},
  pages={3445--3451},
  year={2025},
  organization={IEEE}
}

@inproceedings{xing2025goalflow,
  title={Goalflow: Goal-driven flow matching for multimodal trajectories generation in end-to-end autonomous driving},
  author={Xing, Zebin and Zhang, Xingyu and Hu, Yang and Jiang, Bo and He, Tong and Zhang, Qian and Long, Xiaoxiao and Yin, Wei},
  booktitle={Proceedings of the Computer Vision and Pattern Recognition Conference},
  pages={1602--1611},
  year={2025}
}

@article{chen2024vadv2,
  title={Vadv2: End-to-end vectorized autonomous driving via probabilistic planning},
  author={Chen, Shaoyu and Jiang, Bo and Gao, Hao and Liao, Bencheng and Xu, Qing and Zhang, Qian and Huang, Chang and Liu, Wenyu and Wang, Xinggang},
  journal={arXiv preprint arXiv:2402.13243},
  year={2024}
}

@article{sima2025centaur,
  title={Centaur: Robust end-to-end autonomous driving with test-time training},
  author={Sima, Chonghao and Chitta, Kashyap and Yu, Zhiding and Lan, Shiyi and Luo, Ping and Geiger, Andreas and Li, Hongyang and Alvarez, Jose M},
  journal={arXiv preprint arXiv:2503.11650},
  year={2025}
}

@article{li2024hydra,
  title={Hydra-mdp: End-to-end multimodal planning with multi-target hydra-distillation},
  author={Li, Zhenxin and Li, Kailin and Wang, Shihao and Lan, Shiyi and Yu, Zhiding and Ji, Yishen and Li, Zhiqi and Zhu, Ziyue and Kautz, Jan and Wu, Zuxuan and others},
  journal={arXiv preprint arXiv:2406.06978},
  year={2024}
}

@InProceedings{Li_2025_ICCV,
    author    = {Li, Yingyan and Wang, Yuqi and Liu, Yang and He, Jiawei and Fan, Lue and Zhang, Zhaoxiang},
    title     = {End-to-End Driving with Online Trajectory Evaluation via BEV World Model},
    booktitle = {Proceedings of the IEEE/CVF International Conference on Computer Vision (ICCV)},
    month     = {October},
    year      = {2025},
    pages     = {27137-27146}
}

@InProceedings{Liao_2025_CVPR,
    author    = {Liao, Bencheng and Chen, Shaoyu and Yin, Haoran and Jiang, Bo and Wang, Cheng and Yan, Sixu and Zhang, Xinbang and Li, Xiangyu and Zhang, Ying and Zhang, Qian and Wang, Xinggang},
    title     = {DiffusionDrive: Truncated Diffusion Model for End-to-End Autonomous Driving},
    booktitle = {Proceedings of the IEEE/CVF Conference on Computer Vision and Pattern Recognition (CVPR)},
    month     = {June},
    year      = {2025},
    pages     = {12037-12047}
}

@article{zou2025diffusiondrivev2,
  title={DiffusionDriveV2: Reinforcement Learning-Constrained Truncated Diffusion Modeling in End-to-End Autonomous Driving},
  author={Zou, Jialv and Chen, Shaoyu and Liao, Bencheng and Zheng, Zhiyu and Song, Yuehao and Zhang, Lefei and Zhang, Qian and Liu, Wenyu and Wang, Xinggang},
  journal={arXiv preprint arXiv:2512.07745},
  year={2025}
}

@article{guo2025ipad,
  title={iPad: Iterative Proposal-centric End-to-End Autonomous Driving},
  author={Guo, Ke and Liu, Haochen and Wu, Xiaojun and Pan, Jia and Lv, Chen},
  journal={arXiv preprint arXiv:2505.15111},
  year={2025}
}

@article{chi2025diffusion,
  title={Diffusion policy: Visuomotor policy learning via action diffusion},
  author={Chi, Cheng and Xu, Zhenjia and Feng, Siyuan and Cousineau, Eric and Du, Yilun and Burchfiel, Benjamin and Tedrake, Russ and Song, Shuran},
  journal={The International Journal of Robotics Research},
  volume={44},
  number={10-11},
  pages={1684--1704},
  year={2025},
  publisher={Sage Publications Sage UK: London, England}
}

@misc{jiang2025transdiffuserdiversetrajectorygeneration,
      title={TransDiffuser: Diverse Trajectory Generation with Decorrelated Multi-modal Representation for End-to-end Autonomous Driving}, 
      author={Xuefeng Jiang and Yuan Ma and Pengxiang Li and Leimeng Xu and Xin Wen and Kun Zhan and Zhongpu Xia and Peng Jia and Xianpeng Lang and Sheng Sun},
      year={2025},
      eprint={2505.09315},
      archivePrefix={arXiv},
      primaryClass={cs.RO},
      url={https://arxiv.org/abs/2505.09315}, 
}

@article{dauner2024navsim,
  title={Navsim: Data-driven non-reactive autonomous vehicle simulation and benchmarking},
  author={Dauner, Daniel and Hallgarten, Marcel and Li, Tianyu and Weng, Xinshuo and Huang, Zhiyu and Yang, Zetong and Li, Hongyang and Gilitschenski, Igor and Ivanovic, Boris and Pavone, Marco and others},
  journal={Advances in Neural Information Processing Systems},
  volume={37},
  pages={28706--28719},
  year={2024}
}

@article{cao2025pseudo,
  title={Pseudo-simulation for autonomous driving},
  author={Cao, Wei and Hallgarten, Marcel and Li, Tianyu and Dauner, Daniel and Gu, Xunjiang and Wang, Caojun and Miron, Yakov and Aiello, Marco and Li, Hongyang and Gilitschenski, Igor and others},
  journal={arXiv preprint arXiv:2506.04218},
  year={2025}
}

@inproceedings{hu2023_uniad,
 title={Planning-oriented Autonomous Driving}, 
 author={Yihan Hu and Jiazhi Yang and Li Chen and Keyu Li and Chonghao Sima and Xizhou Zhu and Siqi Chai and Senyao Du and Tianwei Lin and Wenhai Wang and Lewei Lu and Xiaosong Jia and Qiang Liu and Jifeng Dai and Yu Qiao and Hongyang Li},
 booktitle={Proceedings of the IEEE/CVF Conference on Computer Vision and Pattern Recognition},
 year={2023},
}

@inproceedings{jiang2023vad,
  title={Vad: Vectorized scene representation for efficient autonomous driving},
  author={Jiang, Bo and Chen, Shaoyu and Xu, Qing and Liao, Bencheng and Chen, Jiajie and Zhou, Helong and Zhang, Qian and Liu, Wenyu and Huang, Chang and Wang, Xinggang},
  booktitle={Proceedings of the IEEE/CVF International Conference on Computer Vision},
  pages={8340--8350},
  year={2023}
}

@inproceedings{
shang2025drivedpo,
title={Drive{DPO}: Policy Learning via Safety {DPO} For End-to-End Autonomous Driving},
author={ShuYao Shang and Yuntao Chen and Yuqi Wang and Yingyan Li and Zhaoxiang Zhang},
booktitle={The Thirty-ninth Annual Conference on Neural Information Processing Systems},
year={2025},
url={https://openreview.net/forum?id=eIf9GNcA5n}
}

@article{li2025ztrs,
  title={Ztrs: Zero-imitation end-to-end autonomous driving with trajectory scoring},
  author={Li, Zhenxin and Yao, Wenhao and Wang, Zi and Sun, Xinglong and Chen, Jingde and Chang, Nadine and Shen, Maying and Song, Jingyu and Wu, Zuxuan and Lan, Shiyi and others},
  journal={arXiv preprint arXiv:2510.24108},
  year={2025}
}

@article{chen2024end,
  title={End-to-end autonomous driving: Challenges and frontiers},
  author={Chen, Li and Wu, Penghao and Chitta, Kashyap and Jaeger, Bernhard and Geiger, Andreas and Li, Hongyang},
  journal={IEEE Transactions on Pattern Analysis and Machine Intelligence},
  year={2024},
  publisher={IEEE}
}

@article{zheng2025resad,
  title={ResAD: Normalized Residual Trajectory Modeling for End-to-End Autonomous Driving},
  author={Zheng, Zhiyu and Chen, Shaoyu and Yin, Haoran and Zhang, Xinbang and Zou, Jialv and Wang, Xinggang and Zhang, Qian and Zhang, Lefei},
  journal={arXiv preprint arXiv:2510.08562},
  year={2025}
}

@article{yin2025diffrefiner,
  title={DiffRefiner: Coarse to Fine Trajectory Planning via Diffusion Refinement with Semantic Interaction for End to End Autonomous Driving},
  author={Yin, Liuhan and Ju, Runkun and Guo, Guodong and Cheng, Erkang},
  journal={arXiv preprint arXiv:2511.17150},
  year={2025}
}

@inproceedings{
song2021denoising,
title={Denoising Diffusion Implicit Models},
author={Jiaming Song and Chenlin Meng and Stefano Ermon},
booktitle={International Conference on Learning Representations},
year={2021},
url={https://openreview.net/forum?id=St1giarCHLP}
}

@inproceedings{
li2025enhancing,
title={Enhancing End-to-End Autonomous Driving with Latent World Model},
author={Yingyan Li and Lue Fan and Jiawei He and Yuqi Wang and Yuntao Chen and Zhaoxiang Zhang and Tieniu Tan},
booktitle={The Thirteenth International Conference on Learning Representations},
year={2025},
url={https://openreview.net/forum?id=fd2u60ryG0}
}

@article{yao2025drivesuprim,
  title={DriveSuprim: Towards Precise Trajectory Selection for End-to-End Planning},
  author={Yao, Wenhao and Li, Zhenxin and Lan, Shiyi and Wang, Zi and Sun, Xinglong and Alvarez, Jose M and Wu, Zuxuan},
  journal={arXiv preprint arXiv:2506.06659},
  year={2025}
}

@inproceedings{zhou2023query,
  title={Query-centric trajectory prediction},
  author={Zhou, Zikang and Wang, Jianping and Li, Yung-Hui and Huang, Yu-Kai},
  booktitle={Proceedings of the IEEE/CVF conference on computer vision and pattern recognition},
  pages={17863--17873},
  year={2023}
}

@inproceedings{weng2024drive,
  title={Para-drive: Parallelized architecture for real-time autonomous driving},
  author={Weng, Xinshuo and Ivanovic, Boris and Wang, Yan and Wang, Yue and Pavone, Marco},
  booktitle={Proceedings of the IEEE/CVF Conference on Computer Vision and Pattern Recognition},
  pages={15449--15458},
  year={2024}
}

@article{yuan2024drama,
  title={Drama: An efficient end-to-end motion planner for autonomous driving with mamba},
  author={Yuan, Chengran and Zhang, Zhanqi and Sun, Jiawei and Sun, Shuo and Huang, Zefan and Lee, Christina Dao Wen and Li, Dongen and Han, Yuhang and Wong, Anthony and Tee, Keng Peng and others},
  journal={arXiv preprint arXiv:2408.03601},
  year={2024}
}

@article{feng2025artemis,
  title={Artemis: Autoregressive end-to-end trajectory planning with mixture of experts for autonomous driving},
  author={Feng, Renju and Xi, Ning and Chu, Duanfeng and Wang, Rukang and Deng, Zejian and Wang, Anzheng and Lu, Liping and Wang, Jinxiang and Huang, Yanjun},
  journal={arXiv preprint arXiv:2504.19580},
  year={2025}
}

@article{song2025breaking,
  title={Breaking imitation bottlenecks: Reinforced diffusion powers diverse trajectory generation},
  author={Song, Ziying and Liu, Lin and Pan, Hongyu and Liao, Bencheng and Guo, Mingzhe and Yang, Lei and Zhang, Yongchang and Xu, Shaoqing and Jia, Caiyan and Luo, Yadan},
  journal={arXiv e-prints},
  pages={arXiv--2507},
  year={2025}
}

@article{caesar2021nuplan,
  title={nuplan: A closed-loop ml-based planning benchmark for autonomous vehicles},
  author={Caesar, Holger and Kabzan, Juraj and Tan, Kok Seang and Fong, Whye Kit and Wolff, Eric and Lang, Alex and Fletcher, Luke and Beijbom, Oscar and Omari, Sammy},
  journal={arXiv preprint arXiv:2106.11810},
  year={2021}
}

@inproceedings{contributors2023openscene,
  title={Openscene: The largest up-to-date 3d occupancy prediction benchmark in autonomous driving},
  author={Contributors, OpenScene},
  booktitle={Proceedings of the Conference on Computer Vision and Pattern Recognition, Vancouver, Canada},
  pages={18--22},
  year={2023}
}

@inproceedings{he2016deep,
  title={Deep residual learning for image recognition},
  author={He, Kaiming and Zhang, Xiangyu and Ren, Shaoqing and Sun, Jian},
  booktitle={Proceedings of the IEEE conference on computer vision and pattern recognition},
  pages={770--778},
  year={2016}
}
\bibliographystyle{icml2026}

\newpage
\appendix
\onecolumn

\section{Additional Method Details}
\label{app:method_details}

\subsection{HATNA Details}
\label{app:hatna_details}

Recall that diffusion is performed in the position-sequence space $\mathbf{p}=\{(x_i,y_i)\}_{i=1}^{n}$.
In the forward process, we replace the base noise $\epsilon$ in~\cref{eq:forward_process} with an adapted noise $\epsilon'$:
\begin{equation}
\epsilon'=\mathrm{HATNA}(\epsilon),\qquad 
\mathbf{p}_t=\sqrt{\bar{\alpha}_t}\mathbf{p}_0+\sqrt{1-\bar{\alpha}_t}\,\epsilon'.
\end{equation}
HATNA consists of two steps: temporal low-pass smoothing along the horizon dimension and horizon-aware scale modulation.
Below we detail both components.

\paragraph{Temporal smoothing.}
Let $\epsilon\in\mathbb{R}^{B\times M\times n\times D}$ denote the injected noise for a batch of $B$ scenes,
$M$ candidates per scene, $n$ waypoints, and $D$ state dimensions.
We apply a 1D low-pass filter along the waypoint (time) axis to suppress ``zig-zag'' artifacts:
\begin{equation}
\mathcal{S}(\epsilon)_{b,m,i,d}=\sum_{j=-r}^{r} g_j\,\epsilon_{b,m,i+j,d},
\label{eq:hatna_smoothing}
\end{equation}
where $\{g_j\}_{j=-r}^{r}$ is a symmetric, normalized Gaussian kernel (i.e., $\sum_j g_j=1$), and boundary indices are handled by standard padding.
This operation is applied independently for each candidate and each dimension, acting as a temporal low-pass operator.

\paragraph{Horizon-aware scale modulation.}
After smoothing, we modulate the noise magnitude as a function of the waypoint index to reflect increasing uncertainty toward the far horizon.
We first define a monotone base scale profile $\mathbf{s}\in\mathbb{R}^{n}$:
\begin{equation}
s_i = \left(\frac{i}{n-1}+\varepsilon\right)^{\alpha},
\qquad i=0,\ldots,n-1,
\label{eq:hatna_base_scale}
\end{equation}
where $\alpha>0$ controls how rapidly the scale increases with horizon, and $\varepsilon$ is a small constant to avoid degeneracy at $i=0$.
We optionally introduce a learnable, per-waypoint gain $\mathbf{g}\in\mathbb{R}^{n}$ parameterized in log-space:
\begin{equation}
\tilde{s}_i = s_i \cdot \exp(g_i).
\label{eq:hatna_gain}
\end{equation}
Finally, the adapted noise is obtained by element-wise scaling:
\begin{equation}
\epsilon'_{b,m,i,d} = \mathcal{S}(\epsilon)_{b,m,i,d}\cdot \tilde{s}_i.
\label{eq:hatna_scaling}
\end{equation}
Equivalently, $\mathrm{HATNA}(\epsilon)=\mathcal{S}(\epsilon)\odot \tilde{\mathbf{s}}$, where $\tilde{\mathbf{s}}=\{ \tilde{s}_i \}_{i=0}^{n-1}$ is broadcast to match the noise tensor shape.


\subsection{Supervision for the Latent World Model}
\label{app:lwm_supervision}

We supervise the latent world model (LWM) to predict future scene evolution in the BEV latent space.
Given the current observation encoding $z$ (a BEV feature representation) and a candidate trajectory $\tau$, the LWM predicts future BEV features:
\begin{equation}
\hat{z} = \mathrm{LWM}(z,\tau),
\end{equation}
where $\hat{z}$ denotes the predicted future BEV representation.
To provide dense and structured supervision, we decode $\hat{z}$ into a BEV semantic map $\hat{\mathbf{z}}^{+}$.
The corresponding ground-truth future BEV semantic map $\hat{\mathbf{y}}^{+}$ is obtained from the nuPlan semantic-map simulator.
We supervise semantic prediction using a focal loss:
\begin{equation}
\mathcal{L}_{\mathrm{lwm}} = Focalloss\big(\hat{\mathbf{y}}^{+}, \hat{\mathbf{z}}^{+}\big),
\end{equation}
which encourages the world model to capture rare yet critical semantics and improves long-horizon consistency in the latent rollout.


\subsection{Auxiliary Supervision Signals}
\label{app:aux_losses}

Besides world-model supervision, we adopt auxiliary losses to stabilize representation learning and improve downstream planning quality.
First, we apply a standard BEV semantic supervision on the current frame, predicting the present semantic map $\mathbf{z}^{+}$ from the current BEV feature and supervising it with the corresponding ground-truth semantic map $\mathbf{y}^{+}$:
\begin{equation}
\mathcal{L}_{\mathrm{bev}}=Focalloss\big(\mathbf{y}^{+},\mathbf{z}^{+}\big).
\end{equation}
Second, we include agent-related supervision signals derived from annotated agents in the scene (detection targets in BEV), which regularize the BEV features to retain interaction-relevant dynamics:
\begin{equation}
\mathcal{L}_{\mathrm{agent}}=Focalloss(\text{agent predictions},\text{agent targets}).
\end{equation}
In practice, these auxiliary terms are combined with the main objectives in~\cref{sec:learning_objectives} to facilitate stable multi-task learning.


\section{Additional Experimental Details}
\label{app:exp_details}

This section reports implementation details omitted in the main paper, including hyperparameters for HATNA, loss weights, and score aggregation weights.

\subsection{HATNA Hyperparameters}
\label{app:hatna_hparams}

We specify the temporal smoothing kernel size and the horizon scaling profile parameters used in~\cref{eq:hatna_smoothing,eq:hatna_base_scale,eq:hatna_gain}.
Unless otherwise stated, we use a fixed Gaussian smoothing kernel and enable the optional per-waypoint gain for additional flexibility.
\begin{table}[t]
  \caption{\textbf{Additional hyperparameters and weights.} Values in the weight blocks are placeholders (set to 1) unless otherwise specified.}
  \label{tab:app_hparams_weights}
  \centering
  \vspace{0.03in}
  \begin{small}
  \setlength{\tabcolsep}{8pt}
  \renewcommand{\arraystretch}{1.05}
  \begin{tabular}{clc}
    \toprule
    Block & Parameter & Value \\
    \midrule
    \multirow{9}{*}{HATNA} 
      & Number of waypoints $n$ & 8 \\
      & Noise dimension $D$ & 2 \\
      & Smoothing kernel type & Gaussian \\
      & Smoothing kernel size $K$ & 5 \\
      & Horizon scale profile & $(\tfrac{i}{n-1}+\varepsilon)^{\alpha}$ \\
      & Exponent $\alpha$ & 1.0 \\
      & Learnable per-waypoint gain $\exp(g_i)$ & enabled \\
      & Small constant $\varepsilon$ & $10^{-6}$ \\
    \midrule
    \multirow{6}{*}{Loss weights}
      & Trajectory WTA loss weight $\lambda_{traj}$ & 4 \\
      & Imitation scoring loss weight $\lambda_{im}$ & 0.01 \\
      & Simulation scoring loss weight $\lambda_{sim}$ & 0.1 \\
      & World-model semantic loss weight $\lambda_{lwm}$ & 0.1 \\
      & Current BEV semantic loss weight $\lambda_{bev}$ & 10 \\
      & Agent supervision loss weight $\lambda_{agent}$ & 0.1 \\
    \midrule
    \multirow{6}{*}{Score weights}
      & Imitation score weight $w_{\mathrm{im}}$ & 0.05 \\
      & Simulation (NC) score weight $w_{\mathrm{NC}}$ & 0.5 \\
      & Simulation (DAC) score weight $w_{\mathrm{DAC}}$ & 0.5 \\
      & Simulation (EP) score weight $w_{\mathrm{EP}}$ & 1 \\
      & Simulation (TTC) score weight $w_{\mathrm{TTC}}$ & 1 \\
      & Simulation (Comf.) score weight $w_{\mathrm{Comf}}$ & 1 \\
    \bottomrule
  \end{tabular}
  \end{small}
  \vskip -0.08in
\end{table}

\subsection{Loss Weights}
\label{app:loss_weights}

We report the weights for each loss term in the overall objective, including
$\lambda_{traj}$, $\lambda_{im}$, and $\lambda_{sim}$ in~\cref{sec:learning_objectives},
as well as auxiliary weights for $\mathcal{L}_{\mathrm{lwm}}$, $\mathcal{L}_{\mathrm{bev}}$, and $\mathcal{L}_{\mathrm{agent}}$ in~\cref{app:lwm_supervision,app:aux_losses}.

\subsection{Score Aggregation Weights}
\label{app:score_weights}

We provide the fixed weights $\mathbf{w}$ used in the decision rule to aggregate the multi-term score vector into an overall score:
\begin{equation}
S(\tau;z)=\mathbf{w}^{\top}\mathbf{s}(\tau;z),
\end{equation}
and discuss how the weights are set consistently across experiments.
\section{Additional Analysis}
\label{app:analysis}

\subsection{Quantitative Comparison: Regression vs.\ Diffusion Refinement}
\label{app:analysis_quant}

We compare regression-based refinement and diffusion-based refinement using the Average Displacement Error (ADE).
Given a predicted trajectory $\boldsymbol{\tau}^{pred}$ and the expert trajectory $\boldsymbol{\tau}^{gt}$, ADE is defined as
\begin{equation}
\mathrm{ADE}(\boldsymbol{\tau}^{pred}, \boldsymbol{\tau}^{gt})
= \frac{1}{|\mathcal{T}|}\sum_{t\in\mathcal{T}}
\left\lVert \mathbf{p}^{pred}_{t} - \mathbf{p}^{gt}_{t} \right\rVert_{2},
\end{equation}
where $\mathbf{p}_t=(x_t,y_t)$ denotes the 2D position at timestep $t$, and $\mathcal{T}$ is the set of overlapping timesteps between prediction and ground truth.
Table~\ref{tab:ade_refinement} reports the ADE for both refiners. Diffusion-based refinement achieves a lower ADE, indicating more accurate scene-adapted corrections.

\begin{table}[t]
  \caption{\textbf{ADE of different refinement strategies.} Lower is better.}
  \label{tab:ade_refinement}
  \centering
  \vspace{0.03in}
  \begin{small}
  \setlength{\tabcolsep}{10pt}
  \renewcommand{\arraystretch}{1.05}
  \begin{tabular}{lc}
    \toprule
    Refinement strategy & ADE $\downarrow$ \\
    \midrule
    Regression-based refinement & 1.46 \\
    Diffusion-based refinement & \textbf{1.05} \\
    \bottomrule
  \end{tabular}
  \end{small}
  \vskip -0.08in
\end{table}

\subsection{Qualitative Analysis in Highly Interactive Scenarios}
\label{app:analysis_qual}

\begin{figure}[t]
  \centering
  \includegraphics[width=\columnwidth]{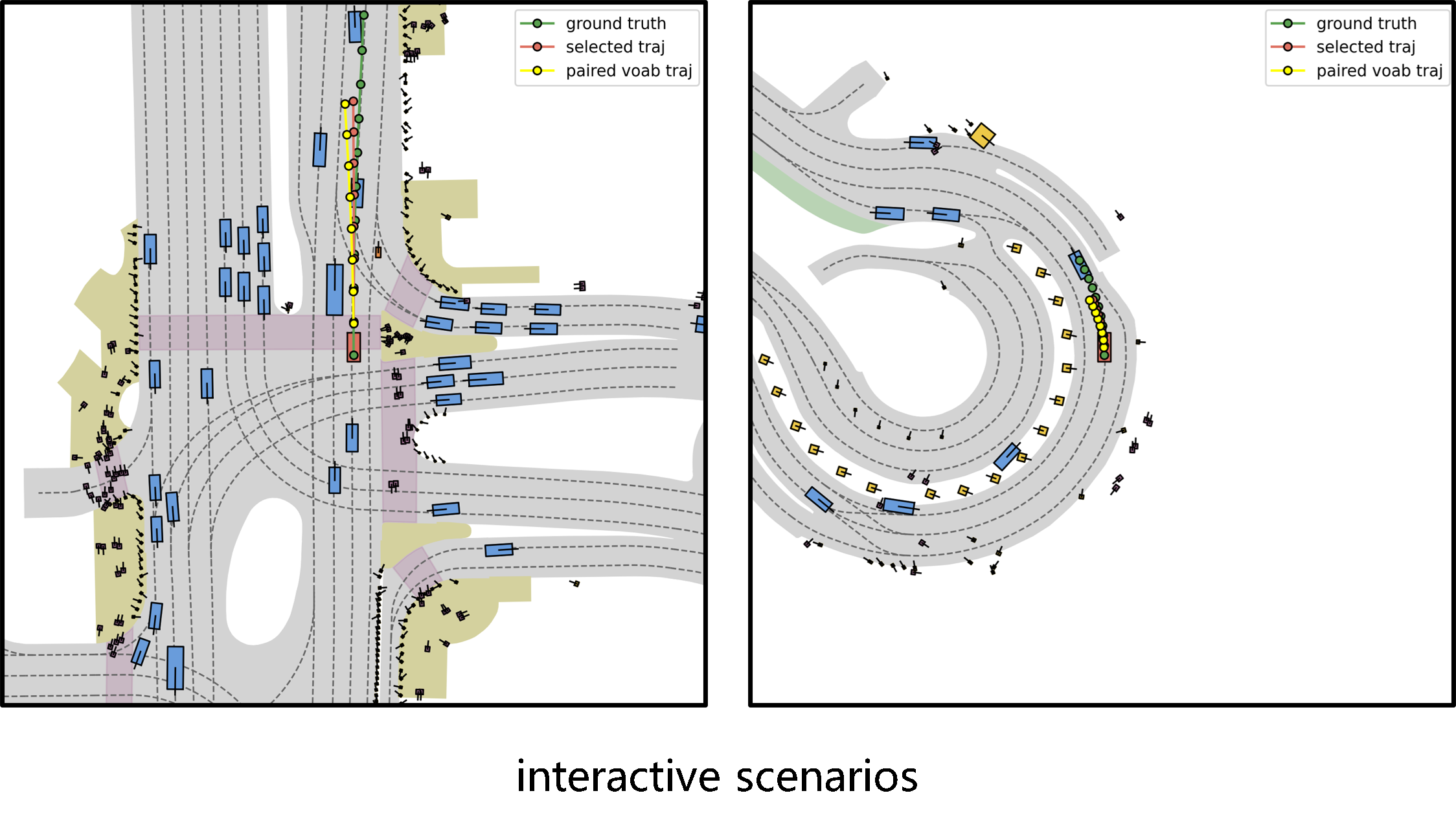}
  \caption{\textbf{Qualitative comparison in highly interactive scenarios.}
  Diffusion-refined candidates better align with interaction-induced constraints than static vocabulary trajectories.}
  \label{fig:qual_interactive}
  \vskip -0.12in
\end{figure}

We further provide qualitative examples in scenarios that require strong interaction with the environment.
In such cases, static vocabulary trajectories may fail to capture fine-grained constraints induced by dynamic agents, yielding conservative or misaligned motions.
In contrast, diffusion-refined trajectories better incorporate context and interaction cues, producing candidates that align more effectively with the scene constraints.
Figure~\ref{fig:qual_interactive} illustrates representative cases where diffusion-refined candidates outperform vocabulary trajectories under high interaction demands.

\end{document}